\documentclass{article}

\usepackage{arxiv}

\usepackage[utf8]{inputenc} 
\usepackage[T1]{fontenc}    
\usepackage{hyperref}       
\usepackage{url}            
\usepackage{booktabs}       
\usepackage{amsfonts}       
\usepackage{nicefrac}       
\usepackage{microtype}      
\usepackage{lipsum}
\usepackage{graphicx}
\usepackage[numbers]{natbib}

\usepackage{verbatim}
\usepackage{fancyvrb}
\usepackage{xcolor}
\usepackage{bm}
\usepackage{listings}

\usepackage[frozencache=true]{minted}

\title{MLJ: A Julia package for composable Machine Learning}

\author{
 Anthony D. Blaom \\
  University of Auckland, Alan Turing Institute, New Zealand eScience Infrastructure\\
  \texttt{anthony.blaom@gmail.com} \\
  \And
 Franz Kiraly \\
  University College London, Alan Turing Institute\\
  \texttt{fkiraly@turing.ac.uk} \\
  \And
 Thibaut Lienart \\
  Alan Turing Institute\\
  \texttt{tlienart@turing.ac.uk} \\
  \And
 Yiannis Simillides \\
  Imperial College London\\
  \texttt{ysimillides@imperial.ac.uk} \\
  \And
 Diego Arenas \\
  University of St Andrews\\
  \texttt{da60@st-andrews.ac.uk} \\
  \And
 Sebastian J. Vollmer \\
  Alan Turing Institute, University of Warwick\\
  \texttt{svollmer@turing.ac.uk} \\
}

\begin{document}
\maketitle

\begin{abstract}
MLJ (Machine Learning in Julia) is an open source software package
providing a common interface for interacting with machine learning
models written in Julia and other languages. It provides tools and
meta-algorithms for selecting, tuning, evaluating, composing and
comparing those models, with a focus on flexible model composition. In
this design overview we detail chief novelties of the framework,
together with the clear benefits of Julia over the dominant
multi-language alternatives.
\end{abstract}


\section{Introduction}

Statistical modeling, and the building of complex modeling pipelines,
is a cornerstone of modern data science. Most experienced data
scientists rely on high-level open source modeling toolboxes - such as
sckit-learn \cite{Pedregosa2001}; \cite{Buitinck2013} (Python); Weka
\cite{Holmes1994} (Java); mlr \cite{BischlEtal2016} and caret
\cite{Kuhn2008} (R) - for quick blueprinting, testing, and creation of
deployment-ready models. They do this by providing a common interface
to atomic components, from an ever-growing model zoo, and by providing
the means to incorporate these into complex work-flows. Practitioners
are wanting to build increasingly sophisticated composite models, as
exemplified in the strategies of top contestants in machine learning
competitions such as Kaggle.

MLJ (Machine Learning in Julia) \cite{MLJdocs} is a toolbox written in
Julia that provides a common interface and meta-algorithms for
selecting, tuning, evaluating, composing and comparing machine model
implementations written in Julia and other languages. More broadly,
the MLJ project hopes to bring cohesion and focus to a number of
emerging and existing, but previously disconnected, machine learning
algorithms and tools of high quality, written in Julia. A welcome
corollary of this activity will be increased cohesion and synergy
within the talent-rich communities developing these tools.

In addition to other novelties outlined below, MLJ aims to provide
first-in-class model composition capabilities. Guiding goals of the
MLJ project have been usability, interoperability, extensibility, code
transparency, and reproducibility.

\subsection{Why Julia?}

Nowadays, even technically competent users of scientific software will
prototype solutions using a high-level language such as python, R, or
MATLAB. However, to achieve satisfactory performance, such code
typically wraps performance critical algorithms written in a second
low-level language, such as C or FORTRAN. Through its use of an
extensible, hierarchical system of abstract types, just-in-time
compilation, and by replacing object-orientation with multiple
dispatch, Julia solves the ubiquitous "two language problem"
\cite{BezansonEtal2017}. With less technical programming knowledge,
experts in a domain of application can get under the hood of machine
learning software to broaden its applicability, and innovation can be
accelerated through a dramatically reduced software development cycle.

As an example of the productivity boost provided by the
single-language paradigm, we cite the DifferentialEquations.jl package
\cite{RackauckasNie2017}, which, in a few short years of development
by a small team of domain experts, became the best package in its
class \cite{Rackauckas2017}.

Another major advantage of a single-language solution is the ability
to automatically differentiate (AD) functions from their code
representations. The Flux.jl package \cite{Innes2018}, for example,
already makes use of AD to allow unparalleled flexibility in neural
network design.

As a new language, Julia is high-performance computing-ready, and its
superlative meta-programming features allow developers to create
domain-specific syntax for user interaction.

\subsection{Novelties}

\paragraph{Composability} In line with current trends in "auto-ML",
MLJ's design is largely predicated on the importance of model
composability. Composite models share all the behaviour of regular
models, constructed using a new flexible "learning networks"
syntax. Unlike the toolboxes cited above, MLJ's composition syntax is
flexible enough to define stacked models, with out-of-sample
predictions for the base learners, as well as more routine linear
pipelines, which can include target transformations that are
learned. As in mlr, hyper-parameter tuning is implemented as a model
wrapper.

\paragraph{A unified approach to probabilistic predictions} In MLJ,
probabilistic prediction is treated as a first class feature,
leveraging Julia's type system. In particular, unnecessary
case-distinctions, and ambiguous conventions regarding the
representation of probabilities, are avoided.

\paragraph{Scientific types} To help users focus less on data
representation (e.g., \texttt{Float32}, \texttt{DataFrame}) and more
on the intended \textit{purpose} or \textit{interpretation} of data,
MLJ articulates model data requirements using \textit{scientific
  types} \cite{ScientificTypes}, such as "continuous", "ordered
factor" or "table".

\paragraph{Connecting models directly to arbitrary data containers} A
user can connect models directly to tabular data in a manifold of
in-memory and out-of-memory formats by using a universal table
interface provided by the Tables.jl package \cite{Quinn}.

\paragraph{Finding the right model} A model registry gives the user
access to model metadata without the need to actually load code
defining the model implementation. This metadata includes the model's
data requirements, for example, as well as a load path to enable MLJ
to locate the model interface code. Users can readily match models to
machine learning tasks, facilitating searches for an optimal model, a
search that can be readily automated.

\paragraph{Tracking classes of categorical variables} Finally, with
the help of scientific types and the CategoricalArrays.jl package
\cite{CategoricalArrays}, users are guided to create safe
representations of categorical data, in which the complete pool of
possible classes is embedded in the data representation, and
classifiers preserve this information when making predictions. This
avoids a pain-point familiar in frameworks that simply recast
categorical data using integers: evaluating a classifier on the test
target, only to find the test data includes classes not seen in the
training data. Preservation of the original labels for these classes
also facilitates exploratory data analysis and interpretability.

\section{Scientific types}

\begin{figure}
  \centering
  \includegraphics[width=0.95\textwidth]{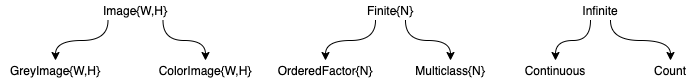}
  \caption{Part of the scientific type hierarchy.}
  \label{fig:fig1}
\end{figure}

A scientific type is an ordinary Julia type (generally without
instances) reserved for indicating how some data should be
interpreted. Some of these types are shown in Figure \ref{fig:fig1}.

To the scientific types, MLJ adds a specific \textit{convention}
specifying a scientific type for every Julia object. The convention is
expressed through a single method \texttt{scitype}.  So, for example,
\texttt{scitype(x)} returns \texttt{Continuous} whenever the type of
\texttt{x} is a subtype of Julia's \texttt{AbstractFloat} type, as in
\texttt{scitype(3.14) == Continuous}. A tabular data structure
satisfying the Tables.jl inteface, will always have type `Table{K}`,
where the type parameter `K` is the union of all column scientific
types. A \texttt{coerce} method recasts machine types to have the
desired scientific type (interpretation), and a \texttt{schema} method
summarizes the machine and scientific types of tabular data.









Since scientific types are also Julia types, Julia's advanced type
system means scientific types can be organized in a type hierarchy.
It is straightforward to check the compatibility of data with a
model's scientific requirements and methods can be dispatched on
scientific type just as they would on ordinary types.




\section{Flexible and compact work-flows for performance evaluation and tuning}

To evaluate the performance of some \texttt{model} object (specifying
the hyper-parameters of some supervised learning algorithm) using some
specified \texttt{resampling} strategy, and measured against some
battery of performance \texttt{measures}, one runs:

\begin{minted}[fontsize=\small]{julia}
    evaluate(model, X, y, resampling=CV(nfolds=6), measures=[L2HingeLoss(), BrierScore()])
\end{minted}

which has (truncated) output:

\begin{minted}[fontsize=\small]{julia}
    ----------------------------------------------------------------------------------------
    | measure                      | measurement | per_fold                                |
    |------------------------------|-------------|-----------------------------------------|
    | L2HingeLoss                  | 1.4         | [0.485, 1.58, 2.06, 1.09, 2.18, 1.03]   |
    | BrierScore{UnivariateFinite} | -0.702      | [-0.24, -0.79, -1.0, -0.54, -1.1, -0.51]|
    ----------------------------------------------------------------------------------------
\end{minted}


As in mlr, hyper-parameter optimization is realized as a model
wrapper, which transforms a base model into a "self-tuning" version of
that model. That is, tuning is is abstractly specified before being
executed. This allows tuning to be integrated into work-flows
(learning networks) in multiple ways. A well-documented tuning
interface \cite{MLJTuning} allows developers to easily extend
available hyper-parameter tuning strategies.

We now give an example of syntax for wrapping a model called
\texttt{forest\_model} in a random search tuning strategy, using
cross-validation, and optimizing the mean square loss. The
\texttt{model} in this case is a composite model with an ordinary
hyper-parameter called \texttt{bagging\_fraction} and a \textit{nested}
hyper-parameter \texttt{atom.n\_subfeatures} (where \texttt{atom} is
another model). The first two lines of code define ranges for these
parameters.

\begin{minted}[fontsize=\small]{julia}
    r1 = range(forest_model, :(atom.n_subfeatures), lower=1, upper=9)
    r2 = range(forest_model, :bagging_fraction, lower=0.4, upper=1.0)
    self_tuning_forest_model = TunedModel(model=forest_model,
                                          tuning=RandomSearch(),
                                          resampling=CV(nfolds=6),
                                          range=[r1, r2],
                                          measure=LPDistLoss(2),
                                          n=25)
\end{minted}

In this random search example, default priors are assigned to each
hyper-parameter, but options exist to customize these. Both resampling
and tuning have options for parallelization; Julia has first class
support for both distributed and multi-threaded parallelism.

\section{A unified approach to probabilistic predictions and their
  evaluation}

MLJ puts probabilistic models and deterministic models on equal
footing. Unlike most most frameworks, a supervised model is either
\textit{probabilistic} - meaning it's \texttt{predict} method returns a
distribution object - \textit{or} it is \textit{deterministic} -
meaning it returns objects of the same scientific type as the training
observations. To use a probabilistic model to make deterministic
predictions one can wrap the model in a pipeline with an appropriate
post-processing function, or use additional \texttt{predict\_mean},
\texttt{predict\_median}, \texttt{predict\_mode} methods to deal with
the common use-cases.

A "distribution" object returned by a probabilistic predictor is one
that can be sampled (using Julia's \texttt{rand} method) and queried
for properties. Where possible the object is in fact a
\texttt{Distribution} object from the Distributions.jl package
\cite{LinEtal2020}, for which an additional \texttt{pdf} method for
evaluating the distribution's probability density or mass function
will be implemented, in addition to \texttt{mode}, \texttt{mean}
and \texttt{median} methods (allowing MLJ's fallbacks for
\texttt{predict\_mean}, etc, to work).

One important distribution \textit{not} provided by Distributions.jl
is a distribution for finite sample spaces with {\em labeled} elements
(called \texttt{UnivariateFinite}) which additionally tracks all
possible classes of the categorical variable it is modeling, and not
just those observed in training data.

By predicting distributions, instead of raw probabilities or
parameters, MLJ avoids a common pain point, namely deciding and
agreeing upon a convention about how these should be represented:
Should a binary classifier predict one probability or two? Are we
using the standard deviation or the variance here? What's the protocol
for deciding the order of (unordered) classes? How should multi-target
predictions be combined?, etc.

A case-in-point concerns performance measures (metrics) for
probabilistic models, such as cross-entropy and Brier loss. All
built-in probabilistic measures provided by MLJ are passed a
distribution in their prediction slot.

For an overview on probabilistic supervised learning we refer to
\cite{Gressmann2018}.

\section{Model interfaces}

In MLJ a \textit{model} is just a struct storing the hyper-parameters
associated with some learning algorithm suggested by the struct name
(e.g., \texttt{DecisionTreeClassifier}) and that is all.  MLJ provides
a basic \textit{model interface}, to be implemented by new machine
learning models, which is functional in style, for simplicity and
maximal flexibility. In addition to a \texttt{fit} and optional
\texttt{update} method, one implements one or more operations, such as
\texttt{predict}, \texttt{transform} and \texttt{inverse\_transform},
acting on the learned parameters returned by \texttt{fit}.

The optional \texttt{update} method allows one to avoid unnecessary
repetition of code execution (warm restart). The three main use-cases
are:

\begin{itemize}

\item \textbf{Iterative models.} If the only change to a random forest
  model is an increase in the number of trees by ten, for example,
  then not all trees need to be retrained; only ten new trees need to
  be trained.

\item \textbf{Data preprocessing.} Avoid overheads associated with
  data preprocessing, such as coercion of data into an
  algorithm-specific type.

\item \textbf{Smart training of composite models.} When tuning a
  simple transformer-predictor pipeline model using a holdout set, for
  example, it is unnecessary to retrain the transformer if only the
  predictor hyper-parameters change. MLJ implements "smart" retraining
  of composite models like this by defining appropriate
  \texttt{update} methods.

\end{itemize}

In the future MLJ will add an \texttt{update\_data} method to support
models that can carry out on-line learning.

Presently, the general MLJ user is encouraged to interact through a
\textit{machine interface} which sits on top of the model
interface. This makes some work-flows more convenient but, more
significantly, introduces a syntax which is more natural in the
context of model composition (see below). A \textit{machine} is a
mutable struct that binds a model to data at construction, as in
\texttt{mach = machine(model, data)}, and which stores learned
parameters after the user calls \texttt{fit!(mach, rows=...)}. To
retrain with new hyper-parameters, the user can mutate \texttt{model}
and repeat the \texttt{fit!} call.

The operations \texttt{predict}, \texttt{transform}, etc are
overloaded for machines, which is how the user typically uses them, as
in the call \texttt{predict(mach, Xnew)}.

\section{Flexible model composition}

Several limitations surrounding model composition are increasingly
evident to users of the dominant machine learning software
platforms. The basic model composition interfaces provided by the
toolboxes mentioned in the Introduction all share one or more of the
following shortcomings, which do not exist in MLJ:

\begin{itemize}

\item Composite models do not inherit all the behavior of ordinary
  models.

\item Composition is limited to linear (non-branching) pipelines.

\item Supervised components in a linear pipeline can only occur at the
  end of the pipeline.

\item Only static (unlearned) target transformations/inverse
  transformations are supported.

\item Hyper-parameters in homogeneous model ensembles cannot be
  coupled.

\item Model stacking, with out-of-sample predictions for base
  learners, cannot be implemented.

\item Hyper-parameters and/or learned parameters of component models
  are not easily inspected or manipulated (in tuning algorithms, for
  example).

\item Composite models cannot implement multiple operations, for
  example, both a \texttt{predict} and \texttt{transform} method (as
  in clustering models) or both a \texttt{transform} and
  \texttt{inverse\_transform} method.

\end{itemize}

We now sketch MLJ's composition API, referring the reader to
\cite{Blaom_I} for technical details, and to the MLJ documentation
\cite{MLJdocs,MLJtutorials} for examples that will clarify how the
composition syntax works in practice.

Note that MLJ also provides "canned" model composition for common use
cases, such as non-branching pipelines and homogeneous ensembles,
which are not discussed further here.

Specifying a new composite model type is in two steps,
\textit{prototyping} and \textit{export}.

\subsection{Prototyping}

In prototyping the user defines a so-called \textit{learning network},
by effectively writing down the same code she would use if composing
the models "by hand". She does this using the machine syntax, with
which she will already be familiar, from the basic
\texttt{fit!}/\texttt{predict} work-flow for single models. There is
no need for the user to provide production training data in this
process. A dummy data set suffices, for the purposes of testing the
learning network as it is built.

\begin{figure}
  \centering
  \mbox{\hspace{0.02\textwidth}}\includegraphics[width=1.05\textwidth]{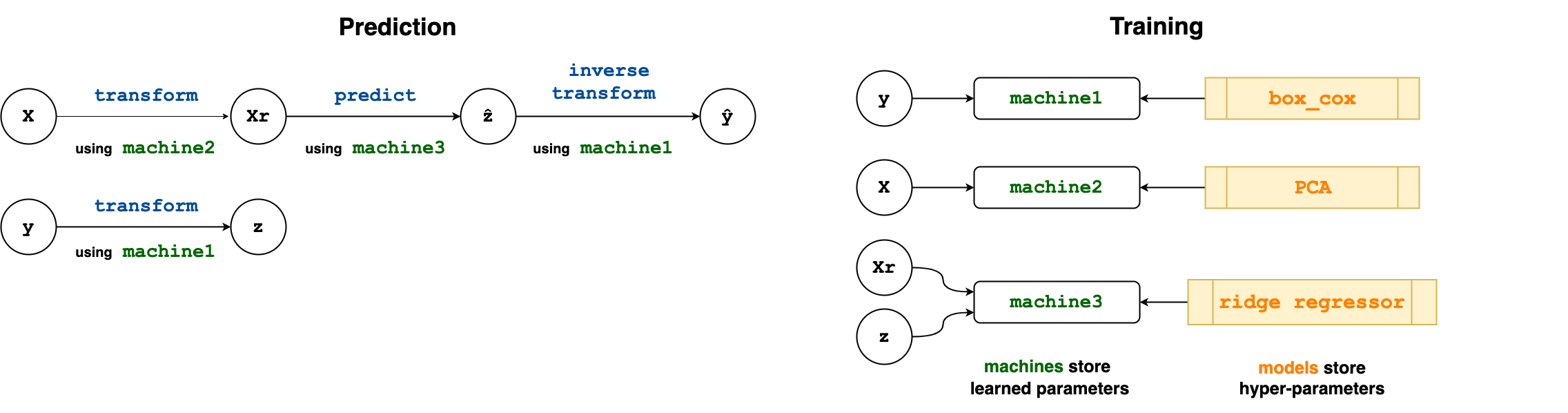}
  \caption{Specifying prediction and training flows in a simple
    learning network. The network shown combines a ridge regressor
    with a learned target transformation (Box Cox).}
  \label{fig:fig2}
\end{figure}

The left panel of Figure \ref{fig:fig2} illustrates a simple
learning network in which a continuous target \texttt{y} is
"normalized" using a learned Box Cox transformation, producing
\texttt{z}, while PCA dimension reduction is applied to some features
\texttt{X}, to obtain \texttt{Xr}. A Ridge regressor, trained using
data from \texttt{Xr} and \texttt{z}, is then applied to \texttt{Xr}
to make a target prediction \texttt{\^{z}}. To obtain a final
prediction \texttt{\^{y}}, we apply the \textit{inverse} of the Box
Cox transform, learned previously, to \texttt{\^{z}}.

The right "training" panel of the figure shows the three machines
which will store the parameters learned in training - the Box Cox
exponent and shift (\texttt{machine1}), the PCA projection
(\texttt{machine2}) and the ridge model coefficients and intercept
(\texttt{machine3}). The diagram additionally indicates where machines
should look for training data, and where to access model
hyper-parameters (stored in \texttt{box\_cox}, \texttt{PCA} and
\texttt{ridge\_regressor}).

The only syntactic difference between composing "by hand" and building
a learning network is that the training data must be wrapped in
"source nodes" (which can be empty if testing is not required) and the
`fit!` calls can be omitted, as training is now lazy. Each data
"variable" in the manual work-flow is now a node of a directed acyclic
graph encoding the composite model architecture. Nodes are callable,
with a node call triggering lazy evaluation of the \texttt{predict},
\texttt{transform} and other operations in the network. Instead of
calling \texttt{fit!} on every machine, a single call to \texttt{fit!}
on a \textit{node} triggers training of all machines needed to call
that node, in appropriate order. As mentioned earlier, training such a
node is "smart" in the sense that hyper-parameter changes to a model
only trigger retraining of necessary machines. So, for example, there
is no need to retrain the Box Cox transformer in the preceding example
if only the ridge regressor hyper-parameters have changed.

The syntax, then, for specifying the learning network shown in Figure
\ref{fig:fig2} looks like this:

\begin{minted}[fontsize=\small,escapeinside=||,mathescape=true]{julia}
    X = source(X_dummy)        # or just source()
    y = source(y_dummy)        # or just source()

    machine1 = machine(box_cox, y)
    z = transform(machine1, y)

    machine2 = machine(PCA, X)
    Xr = transform(machine2, X)

    machine3 = machine(ridge_regressor, Xr, z)
    |$\hat{\texttt{z}}$| = predict(machine3, Xr)

    ŷ = inverse_transform(machine1, |$\hat \texttt{z}$|)

    fit!(ŷ)  # to test training on the dummy data
    ŷ()      # to test prediction on the dummy data
\end{minted}


Note that the machine syntax is a mechanism allowing for multiple
nodes to point to the same learned parameters of a model, as in the
learned target transformation/inverse transformation above. They also
allow multiple nodes to share the same model (hyper-parameters) as in
homogeneous ensembles. And different nodes can be accessed during
training and "prediction" modes of operation, as in stacking.

\subsection{Export}

In the second step of model composition, the learning network is
"exported" as a new stand-alone composite model type, with the
component models appearing in the learning network becoming default
values for corresponding hyper-parameters of the composite. This new
type (which is unattached to any particular data) can be instantiated
and used just like any other MLJ model (tuned, evaluated, etc). Under
the hood, training such a model builds a learning network, so that
training is "smart". Defining a new composite model type requires
generating and evaluating code, but this is readily implemented using
Julia's meta-programming tools, i.e., executed by the user with a
simple macro call.

\section{Future directions}

There are plans to: (i) grow the number of models; (ii) enhance core
functionality, particularly around hyper-parameter optimization
\cite{MLJTuning}; and (iii) broaden scope, particularly around
probabilistic programming models, time series, sparse data and natural
language processing. A more comprehensive road map is linked from the
MLJ repository \cite{MLJ}.

\section*{Acknowledgments}

We acknowledge valuable conversations with Avik Sengupta, Mike Innes,
mlr author Bernd Bischl, and IQVIA's Yaqub Alwan and Gwyn Jones. Seed
funding for the MLJ project has been provided by the Alan Turing
Institute's Tools, Practices and Systems programme, with special thanks
to Dr James Hethering, its former Programme Director, and Katrina
Payne. Mathematics for Real-World Systems Centre for Doctoral Training
at the University of Warwick provided funding for students exploring
the Julia ML ecosystem, who created an initial proof-of-concept.

\textbf{Code contributors}. D. Aluthge, D. Arenas, E. Barp,
C. Bieganek, A. Blaom, G. Bohner, M. K. Borregaard, D. Buchaca,
V. Churavy, H. Devereux, M. Giordano, J. Hoffimann, T. Lienart,
M. Nook, Z. Nugent, S. Okon, P. Oleśkiewicz, J. Samaroo, A. Shridar,
Y. Simillides, A. Stechemesser, S. Vollmer.

\bibliographystyle{unsrtnat}

\providecommand{\natexlab}[1]{#1}
\providecommand{\url}[1]{\texttt{#1}}
\expandafter\ifx\csname urlstyle\endcsname\relax
  \providecommand{\doi}[1]{doi: #1}\else
  \providecommand{\doi}{doi: \begingroup \urlstyle{rm}\Url}\fi

\end{document}